\documentclass{article}

\usepackage[final, nonatbib]{neurips_2022/neurips_2022}

\usepackage[utf8]{inputenc} \usepackage[T1]{fontenc}    \usepackage{url}            \usepackage{booktabs}       \usepackage{amsfonts}       \usepackage{nicefrac}       \usepackage{microtype}      \usepackage{authblk}

\usepackage{amsmath,amsfonts,amsthm,amssymb}
\usepackage{graphicx}
\usepackage{url}
\usepackage{color}
\usepackage[usenames,dvipsnames,svgnames,table]{xcolor}
\usepackage[colorlinks=true, linkcolor=red, urlcolor=blue, citecolor=blue]{hyperref}
\usepackage[numbers]{natbib}
 \usepackage{booktabs}
\usepackage{algorithm}
\usepackage{algorithmic}
\usepackage{bm}
\usepackage{stackrel}
\usepackage{enumitem}
\usepackage{graphics}
\usepackage{wrapfig}

\newcommand{\Ebb}{\mathbb{E}}
\newcommand{\Hbb}{\mathbb{H}}

\newcommand{\Nbb}{\mathbb{N}}

\newcommand{\Rbb}{\mathbb{R}}

\newcommand{\Xcal}{\mathcal{X}}

\newcommand{\Statespace}{\mathcal{S}}
\newcommand{\Actionspace}{\mathcal{A}}

\DeclareMathOperator{\argmax}{arg \max}
\DeclareMathOperator{\argmin}{arg \min}

\DeclareMathOperator{\EIG}{EIG}

\definecolor{nicegreen}{RGB}{91,226,91}
\newcommand{\add}[1]{#1} \graphicspath{{images/}}

\title{Exploration via Planning for Information about the Optimal Trajectory}

\author{\textbf{
  Viraj Mehta$^1$, Ian Char$^2$, Joseph Abbate$^4$, Rory Conlin$^4$, Mark D. Boyer$^3$, Stefano Ermon$^5$, Jeff Schneider$^1$, Willie Neiswanger$^5$}\\
  \vspace{-1mm}
  $^1$Robotics Institute, $^2$Machine Learning Department, Carnegie Mellon University\\
  $^3$Princeton Plasma Physics Laboratory, $^4$Princeton University\\
  $^5$Computer Science Department, Stanford University\\
  \vspace{2mm}
  \texttt{\{virajm,ichar,schneide\}@cs.cmu.edu, \{jabbate,wconlin\}@princeton.edu, mboyer@pppl.gov, \{ermon,neiswanger\}@cs.stanford.edu}\\
}

\begin{document}

\maketitle

\begin{abstract}
Many potential applications of reinforcement learning (RL) are stymied by the large numbers of samples required to learn an effective policy.
This is especially true when applying RL to real-world control tasks, e.g. in the sciences or robotics, where executing a policy in the environment is costly.
In popular RL algorithms, agents typically explore either by adding stochasticity to a reward-maximizing policy or by attempting to gather maximal information about environment dynamics without taking the given task into account.
In this work, we develop a method that allows us to plan for exploration while taking both the task and the current knowledge about the dynamics into account. 
The key insight to our approach is to plan an action sequence that maximizes the expected information gain about the optimal trajectory for the task at hand.
We demonstrate that our method learns strong policies with \add{2x fewer samples than strong exploration baselines} and \add{200x fewer samples than model free methods} on a diverse set of \add{low-to-medium} dimensional control tasks in both the open-loop and closed-loop control settings.\footnote{Code is available at: \url{https://github.com/fusion-ml/trajectory-information-rl}}

\end{abstract} \section{Introduction}
\label{sec:introduction}
The potential of reinforcement learning (RL) as a general-purpose method of learning solutions to sequential decision making problems is difficult to overstate.
Ideally, RL could allow for agents that learn to accomplish all manner of tasks solely through a given reward function and the agent's experience; however, RL has so far broadly fallen short of this.
Chief among the difficulties in realizing this potential is the fact that typical RL methods in continuous problems require very large numbers of samples to achieve a near-optimal policy.
For many interesting applications of RL this problem is exacerbated by the fact that collecting samples from the true environment can incur huge costs. For example, expensive scientific experiments are needed for tokamak control 
\citep{humphreys2015novel, char2019offline, deepmind_plasma} and design of molecules \citep{zhou2019optimization}, and collecting experience on the road is both costly and runs the risk of car accidents for autonomous vehicles \citep{toromanoff2020end}.

Though model-based methods like \citet{pilco, chua_pets}, and \citet{curi_hucrl} are much more efficient than typical model-free methods, they do not explicitly reason about
the prospective task-relevant information content of future observations.
Moreover, these methods typically require thousands of timesteps of environment dynamics data to
solve reasonably simple Markov decision processes (MDPs).
Though progress has been made in using Thompson sampling \citep{osband_deep_exploration}, entropy bonuses \citep{haarnoja2018soft}, and upper confidence bounds (UCB) \citep{curi_hucrl, ash2022anticoncentrated} to more intelligently explore the state-action space of an MDP,
these methods still do not explicitly reason about how information that the agent gathers will affect the estimated MDP solution.
Furthermore, in continuous state-action settings these methods must make coarse approximations
to be computationally tractable (e.g. bootstrapped Q networks to approxiate a posterior or one-step perturbations for approximate UCB).

Many methods in the vein of \citet{pathak2017curiosity} and \citet{shyam_max} explore directly
to gather new information based on current uncertainty about environment dynamics, but they do not in general specialize the information that they aim to acquire for a particular \emph{task} specified by an initial state distribution and reward function.
We believe that an ideal exploration strategy for sample efficiency should take into account this task, as well as uncertainty about the environment dynamics.

\begin{figure}
    \centering
    \includegraphics[width=\textwidth]{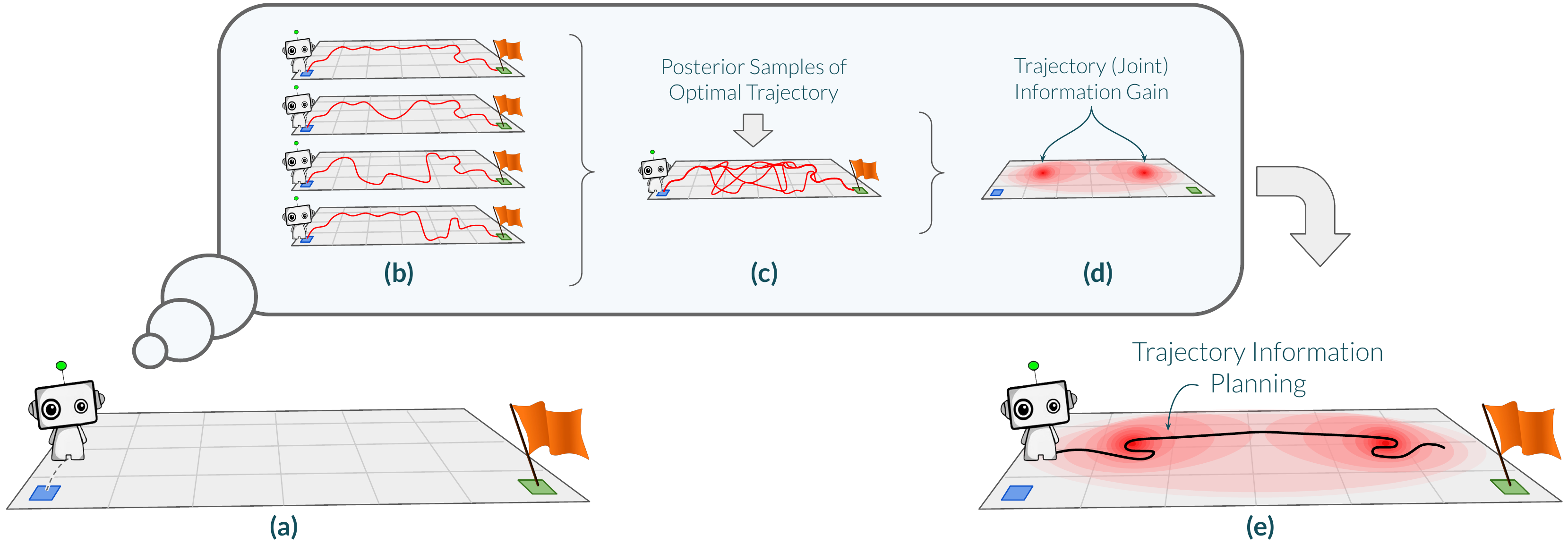}
    \vspace{-3mm}
    \caption{
    \small
    A schematic depiction of Trajectory Information Planning (TIP). Suppose the agent in (a) aims to determine where to explore next from its current state. To do so, in (b) the agent samples dynamics models $T'\sim P(T\mid D)$ from its current posterior and finds approximately optimal trajectories $\tau^* \sim P(\tau^* \mid T')$ for each sample. Then in (c)
it pools these samples of posterior optimal trajectories $\tau^*$. In (d) it constructs a function that gives the joint expected information gain about the optimal trajectory $\tau^*$ given a planned exploration trajectory (i.e. $\EIG_{\tau^*}$ over the set of points visited). Finally, in (e) the agent can plan an action sequence which maximizes this joint expected information gain.}
    \label{fig:tip-cartoon}
    \vspace{-3mm}
\end{figure}

In this work, we start by showing how many methods can be cast as a Bayesian planning problem over a specific cost function. This framework helps illuminate the importance of the cost function and what impact it has on exploration. Viewed in this light, it becomes clear that many previous state-of-the-art methods rely on cost functions that either result in behavior that is too greedy---i.e. the policy tries to maximize returns during exploration---or too exploratory, i.e. the policy is incentivized to explore the environment dynamics and does not consider the task at hand. We therefore present a cost function that balances out these two extremes, \add{by generalizing an information-gain acquisition function introduced in \citet{mehta2022barl} to apply to a set of future hypothetically acquired data}. In particular, our cost function captures the amount of information that would be gained about the \emph{optimal trajectory}, if the agent were to explore by following a particular planned trajectory. As depicted in Figure~\ref{fig:tip-cartoon}, this involves the agent sampling what it would hypothetically do given different realizations of the dynamics, and then planning actions that are informative about those possibilities.

In summary, the contributions of this work are as follows: we develop a novel cost function for exploration that explicitly accounts for both the specific task and uncertainty about environment dynamics, a method for planning which applies the cost function to explore in MDPs with continuous states and actions, and a thorough empirical evaluation of our method across 5 closed-loop and 3 open-loop environments (with a focus on expensive RL tasks in plasma physics) compared against 14 baselines. We find that our proposed method is able to learn policies that perform as well as an agent with access to the ground truth dynamics using \add{half or} fewer samples than comparison methods.

 \vspace{-3mm}
\section{Related Work}
\label{sec:relatedwork}
\vspace{-2mm}
\paragraph{Exploration in Reinforcement Learning}
The most common strategy for exploration in RL is to execute a greedy policy with some form of added stochasticity. The simplest approach,  $\epsilon$-greedy exploration as used in \citet{mnih2013playing}, takes the current action thought to be best with probability $1-\epsilon$ and a random action with probability $\epsilon$. Other methods use added Ornstein-Uhlenbeck action noise \citep{lillicrap2015continuous} to the greedy policy, or entropy bonuses \citep{haarnoja2018soft} to the policy or value function objectives to add noise to a policy which is otherwise optimizing the RL objective.

Tabular RL is often solved by choosing actions based on upper confidence bounds on the value function \citep{ChenUCB, lee2021sunrise}, but explicitly computing and optimizing these bounds in the continuous setting is substantially more challenging. Recent work \citep{curi_hucrl} approximates this method by computing one-step confidence bounds on the dynamics and training a `hallucinated' policy which chooses perturbations within these bound to maximize expected policy performance. Another recent work \citep{ash2022anticoncentrated} uses anti-concentration inequalities to approximate upper confidence bounds in MDPs with discrete actions. 

Thompson sampling (TS) \citep{russo2018tutorial}, which samples a realization of the MDP from the posterior and acts optimally as if the realization was the true model, can be applied for exploration in a model-free manner as in \citep{OsbandBootstrapped} or in a model-based manner as in \cite{strens2000bayesian}.
As the posterior over MDP dynamics or value functions can be high-dimensional and difficult to represent, the performance of TS can be hindered by approximation errors using both Gaussian processes and ensembles of neural networks. \citet{curi_hucrl} recently investigated this and found that this was potentially due to an insufficiently expressive posterior over entire transition functions, implying that it may be quite difficult to solve tasks using sampled models.
Similarly, the posterior over action-value functions in \citet{OsbandBootstrapped} is only roughly approximated by training a bootstrapped ensemble of neural networks.

There is also a rich literature of Bayesian methods for exploration, which are typically computationally expensive and hard to use, though they have attractive theoretical properties. These methods build upon the fundamental idea of the Bayes-adaptive MDP \citep{ross2007bayes}, which we detail in Section \ref{a:bayesian_exploration} alongside a discussion of this literature.

Additionally, a broad set of methods explore to learn about the environment without addressing a specified task. This line of work is characterized by \citet{pathak2017curiosity}, which synthesizes a task-agnostic reward function from model errors. Other techniques include MAX \citep{shyam_max}, which optimizes the information gain about the environment dynamics, Random Network Distillation \citep{burda2018exploration}, which forces the agent to learn about a random neural network across the state space, and Plan2Explore \citep{sekar2020planning}, which prospectively plans to find areas of novelty where the dynamics are uncertain.

\vspace{-2mm}
\paragraph{Bayesian Experimental Design: BOED, BO, BAX, and BARL}
There is a large literature on Bayesian optimal experiment design (BOED) \citep{chaloner1995bayesian} which focuses on efficiently querying a process or function to get maximal information about some quantity of interest.
When the quantity of interest is the location of a function optimum, related strategies have been proposed as the entropy search family of Bayesian optimization (BO) algorithms \citep{hennig2012entropy, hernandez2014predictive}.
Recently, a flexible framework known as Bayesian algorithm execution (BAX) \citep{neiswanger2021bayesian} has been proposed to efficiently estimate properties of expensive black-box functions;
this framework
gives a general procedure for sampling points which are informative about the future execution of a given algorithm that computes the property of interest, thereby allowing the function property to be estimated with far less data.

A subsequent related work \citep{mehta2022barl}, known as Bayesian Active Reinforcement Learning (BARL), uses ideas from BOED and BAX to sample points that are maximally informative about the optimal trajectory in an MDP. However, BARL relies on a setting the authors call Transition Query Reinforcement Learning (TQRL),
which assumes that the environment dynamics can be iteratively queried at an arbitrary sequence of state-action pairs chosen by the agent.
TQRL is thus a highly restrictive setting which is not suitable when data can only be accessed via a trajectory (rollout) of environment dynamics; it typically
relies on an accurate environment simulator of sufficient expense to warrant its use.
Even then, there will likely be differences between simulators and ground truth dynamics for complex systems. Thus, one would ideally like to collect data in real environments.
However, this often requires leaving the TQRL setting, and instead collecting data via trajectories only.

In this paper, we aim to apply
\add{the information-theoretic ideas from BARL but generalize} them to the general MDP setting as well as learn open loop model-based controllers.
The typical method for learning to solve open-loop control problems was demonstrated successfully in \citet{pmlr-v28-tesch13}, where a value function was learned from action sequences to task success. Our method takes a model-based approach to this problem, using similar exploration strategies as Bayesian optimization but benefitting from the more substantial supervision that is typical in dynamics model learning.

 \vspace{-3mm}
\section{Problem Setting}
\label{sec:preliminaries}
\vspace{-2mm}

In this work we deal with finite-horizon discrete-time \emph{Markov decision processes} (MDPs) which consist of a sextuple $\langle \Statespace, \Actionspace, T, r, p_0, H\rangle$
where $\Statespace$ is the state space, $\Actionspace$ is
the action space, $T$ is the transition function
$T: \Statespace \times \Actionspace\to P(\Statespace)$ (using the convention that $P(\Xcal)$ is the set of probability measures over $\Xcal$), $r: \Statespace \times \Actionspace \times \Statespace \to \Rbb$ is a reward function, $p_0(s)$ is a distribution over $\Statespace$ of start states, and $H \in \Nbb$ is the horizon (i.e. the length of an episode).
We always assume $\Statespace, \Actionspace, p_0$, and $H$ are known. We also assume the reward $r$ is known, though our development of the method can easily be generalized to the case where $r$ is unknown.
Our primary object of interest is the transition function $T$, which we learn from data.
We address both open and closed loop control settings.
In the more common closed loop setting, our aim is to find a policy $\pi: \Statespace \to \Actionspace$ that maximizes
Objective \eqref{eq:objective} below.
We will denote trajectories as $\tau \sim p(\tau\mid \pi, T)$ where $\tau = [(s_0, a_0), \dots, (s_{H-1}, a_{H-1}), s_H]$ generated by $s_0 \sim p_0$, $a_i = \pi(s_i)$, and $s_{i + 1} \sim T(s_{i}, a_i)$. We can write the return of a trajectory as $R(\tau) = \sum_{i = 0}^{H-1} r(s_i, a_i, s_{i + 1})$ for the states and actions $s_i, a_i$ that make up $\tau$. The MDP objective can then be written as
\begin{equation}
\label{eq:objective}
    J_T(\pi) = \mathbb{E}_{\tau\sim p(\tau\mid \pi, T)}\left[R(\tau)\right].
\end{equation}

We aim to maximize this objective while minimizing the number of samples from the ground truth transition function $T$ that are required to reach good performance. 
We denote the optimal policy as $\pi^* = \argmax_\pi J_T(\pi)$, which we can assume to be 
deterministic \citep{Sutton+Barto:1998} but not necessarily unique. 
We use $\tau^*$ to denote optimal trajectories, i.e. $\tau^*\sim p(\tau\mid \pi^*, T)$.

Similarly, for the open-loop setting, we assume a fixed start state $s_0$ and aim to find an action sequence $a_0, \dots, a_{H - 1}$ that maximizes the sum of rewards in an episode. We will slightly abuse notation and write $\tau \sim p(\tau\mid a_{0:H-1})$ and $J_T(a_{0:H-1})$ with these actions fixed in place of a reactive policy, and again use $\tau^*$ to refer to the trajectories generated by an optimal action sequence.

We assume in this work that applying planning algorithms like \citep{pinneri2020iCEM} to a dynamics function $T$ will result in a trajectory that approximates $\tau^*$.
We will primarily focus on a Gaussian process (GP) model of the transition dynamics in order to take advantage of its expressive representation of uncertainty and grounded methods for sampling, conditioning, and joint inference. There is substantial prior work using GPs in RL---see Section \ref{a:gp_rl} for a discussion of this literature.
Under this modeling choice, we assume that the dynamics are drawn from a GP prior $P(T)$ (see Section~\ref{a:gp_model} for further details on our GP model) and use $P(T\mid D)$ for the posterior over transition functions given a dynamics dataset of triples $D = \{(s_i, a_i, s'_i)\}$. In this work, unions $D\cup \tau$ or $D\cup S'$ between the dataset $D$ and trajectories $\tau$ or next state predictions $S'$ coerce $\tau$ and $S'$ into triples of dynamics data, prior to the union with the dataset.

 \vspace{-2mm}
\section{Trajectory Information Planning}
\vspace{-1mm}
Our method consists of a generic framework for Bayesian (or approximately Bayesian) model-predictive control and a novel cost function for planning that allows us to explicitly plan to find the maximal amount of new information relevant to our task. In Section~\ref{sec:methods-mpc}, we describe the MPC framework and highlight that many prior methods approximate this framework while using a greedy cost function that corresponds to the future negative expected rewards or a pure exploration cost function that corresponds to future information about the dynamics. Afterwards, in Section~\ref{sec:methods-acqplanning}, we derive our new cost function and describe how it is computed. The overall method we introduce simply applies this planning framework with our new cost function.

\vspace{-1mm}
\subsection{Model-Predictive Control in Bayesian Model-Based RL}
\label{sec:methods-mpc}
\vspace{-1mm}

In this section, we give a formulation of Bayesian planning for control that generalizes ideas from methods such as PILCO \citep{pilco} and PETS \citep{chua_pets}. This formulation highlights these methods' inherently greedy nature and hints at a possible solution. The objective of Bayesian planning is to find the $h$-step action sequence that maximizes the expected future returns under model uncertainty. That is,
\begin{equation}
\label{eq:bayes_planning}
\underset{a_0, \ldots, a_{h-1}}{\textrm{argmin}} \Ebb_{T'\sim P(T\mid D), \tau_e\sim P(\tau\mid s_0=s, a_{0:h-1}, T')}\left[C(\tau_e)\right]
\end{equation}
for some cost function $C$ over trajectories and some start state $s$.
If operating in the open-loop control setting, the agent executes the sequence of actions found without replanning. This procedure can also be extended to closed-loop control via model-predictive control (MPC), which involves re-planning \eqref{eq:bayes_planning} at every state the agent visits and playing the first action from the optimal sequence. Concretely, the MPC policy for our Bayesian setting is as follows:
\begin{equation}
\label{eq:bayes_mpc}
    \pi_{\text{MPC}}(s) = \underset{a_0}{\argmin} \min_{a_1, \dots, a_{h - 1}}\Ebb_{T'\sim P(T\mid D), \tau_e\sim P(\tau\mid s_0=s, a_{0:h-1}, T')}\left[C(\tau_e)\right]
\end{equation}

Whether we do open-loop control or closed-loop control via MPC, the cost function $C$, is integral to how the agent will behave. Prior work has predominantly focused on two types of cost function:
\begin{align}
    \label{eq:cost_functions}
    \underbrace{C_g(\tau) = -R\left(\tau\right)}_{\text{Greedy Exploration}}
&& \underbrace{C_e(\tau) = - \sum_{i=0}^h \Hbb\left[T(s_i, a_i) \mid D \right]}_{\text{Task-Agnostic Exploration}}
\end{align}
Previous works such as \citet{pmlr-v84-kamthe18a} and PETS \citep{chua_pets} use the greedy exploration cost function, $C_g$. This cost function incentivizes trajectories that achieve high rewards over the next $h$ transitions on average. In works that focus on task-agnostic exploration such as \citet{sekar2020planning} and \citet{shyam_max}, the cost function $C_e$ (or similar) is used to encourage the agent to find areas of the state space in which the model is maximally uncertain. Note that \add{we use $\pi_g$ to refer to the greedy policy given by using \eqref{eq:bayes_mpc} with $C_g$.}

The optimization problem in \eqref{eq:bayes_mpc} is typically approximately solved in one of three ways: \citet{pilco} and \citet{curi_hucrl} directly backpropagate through the estimated dynamics and reward functions to find policy parameters that would generate good actions, \citet{janner2019trust} use an actor-critic method trained via rollouts in the model alongside the data collected to find a policy, and \citet{chua_pets} and \citet{mehta2022barl} use the cross-entropy method \citep{de2005tutorial} to find action sequences which directly maximize the reward over the estimated dynamics.
In this work, we use a version of the last method given in \citet{pinneri2020iCEM}, denoted iCEM, to directly find action sequences that optimize the cost function being used. 
We approximate the expectation by playing the actions on multiple samples from the posterior $P(T\mid D)$.
Algorithm~\ref{alg:mpc} gives a formal description of the method and Section \ref{a:planning} provides further details.

\begin{algorithm}
\caption{Bayesian Model-Predictive Control with Cost Function $C$}
\begin{algorithmic}
\STATE \textbf{Inputs:} transition function episode query budget $b$, number of posterior function samples $k$, planning horizon $h$.
\STATE Initialize $D \leftarrow \emptyset$.
\FOR{$i \in [1, \dots, b]$}
\STATE Sample start state $s_0\sim p_0$.
\FOR{$t \in [0, \dots, H-1]$}
\STATE Sample posterior functions $\{T'_\ell\}_{\ell=1}^k \sim P(T'\mid D)$.
\STATE Approximately find $\argmin_{a_0, \dots, a_{h-1}}\sum_{\ell=1}^k\Ebb_{\tau_\ell\sim p(\tau\mid T'_\ell, a_0, \dots, a_{h-1})}\left[C(\tau_\ell)\right]$ via iCEM.
\STATE Execute action $a_0$ by sampling $s_{t + 1} \sim T(s_t, a_0)$.
\STATE Update dataset $D \gets D \cup \{(s_t, a_0, s_{t + 1}\}$.
\ENDFOR
\ENDFOR
\RETURN $\pi_g$ for the posterior $P(T'\mid D)$.
\end{algorithmic}
\label{alg:mpc}
\end{algorithm}

\vspace{-3mm}
\subsection{A Task-Specific Cost Function based on Trajectory Information}
\label{sec:methods-acqplanning}
\vspace{-1mm}
In this work, we aim to explore by choosing actions 
that maximize the conditional expected information gain (EIG) about the optimal trajectory $\tau^*$.
This is the same overall goal as that of \citet{mehta2022barl}, where the $\EIG_{\tau^*}$ acquisition function was introduced for this purpose.
However, in this paper we generalize this acquisition function in order to allow for sequential information collection, and account for the redundant information that could be collected between timesteps. As discussed at length in \citet{osband_deep_exploration}, 
it is essential to reason about how an action taken at the current timestep will affect the possibility of learning something useful in future timesteps.
In other words, exploration must be \emph{deep} and not greedy. Explicit examples are given in \citet{osband_deep_exploration} where the time to find an $\epsilon$-optimal policy in a tabular MDP is exponential in the state size unless exploration can be coordinated over large numbers of timesteps rather than being conducted independently at each action.
As the $\EIG_{\tau^*}$
acquisition function is only defined over a single state-action pair and mutual information is submodular, we cannot naively use the acquisition function as is (or sum it over many datapoints) to choose actions that lead to good long-term exploration.
This is clear in e.g. navigation tasks, where the nearby points visited over trajectories will provide redundant information about the local environment.

We therefore give a cost function that generalizes $\EIG_{\tau^*}$ by taking a set of points to query and computing the \emph{joint} expected information gain from observing the set. Our cost function is non-Markovian in the state space of the MDP, but it is Markovian in the dataset, which represents a point in the belief space of the agent about the dynamics. Let $\add{\Xcal} = \{x : x \subseteq \Statespace \times \Actionspace, |x| < \infty\}$ be the set of finite subsets of the set of all state-action pairs. Our cost function $C_{\tau^*}: \add{\Xcal} \to \mathbb{R}$ is defined below to be the negative \emph{joint expected information gain} about the optimal trajectory $\tau^*$ for a subset $\add{X} \in \add{\Xcal}$.
In particular, assuming an existing dataset $D$, a set of $h$ query points $\add{X}=\{(s_i, a_i)\}_{i\in [h]}$, and a random set of next states $S' = \{s'_i \sim T(s_i, a_i), i \in [h]\}$,
\begin{equation}
\label{eq:eig_batch}
C_{\tau^*}(\add{X}) =  \Ebb_{S'\sim p(S'\mid \add{X}, D)}\left[\Hbb\left[\tau^*\mid D\cup S'\right]\right] - \Hbb\left[\tau^*\mid D\right].
\end{equation}
This formulation of $C_{\tau^*}$ forces our method to handle the redundant information among queries---it is likely that $I(s_1', \tau^*) + I(s_2', \tau^*) > I(\{s_1', s_2'\}, \tau^*)$ and our method should avoid this overestimation.
However, as written, this function relies on computing entropies on high-dimensional trajectories where the form of the joint distribution of the elements is unknown. 
To tractably estimate this quantity, we use the fact that $C_{\tau^*}(\add{X}) = -I(S', \tau^*) = -I(\tau^*, S')$ for the mutual information $I$.
This allows us to exchange $\tau^*$ and our set of queries so that $\tau^*$ is giving information about the posterior predictive distribution of our set. In other words,
\begin{equation}
C_{\tau^*}(\add{X}) =  \Ebb_{\tau^*\sim p(\tau^*\mid D)}\left[\Hbb\left[S'\mid D\cup\tau^*\right]\right] - \Hbb\left[S'\mid D\right].
\end{equation}

In order to compute the right-hand term, we must take samples $\tau^*_{ij} \sim P(\tau^*\mid D), i=1, \dots, m, j=1, \dots, n$.
To do this, we first sample $m$ start states $s_0^{(i)}$ from $p_0$ (we always set $m=1$ in experiments but derive the procedure in general) and for each start state independently sample $n$ posterior functions $T'_{ij} \sim P(T'\mid D)$ from our posterior over dynamics models.
We then run a planning procedure using iCEM \citep{pinneri2020iCEM} on each of the posterior functions from $s_0^{(i)}$ using $T'_{ij}$ for $T$ (using our assumption that planning can generate approximately optimal trajectories given ground-truth dynamics), giving our sampled $\tau^*_{ij}$.
Formally, we can approximate $C_{\tau^*}$ via Monte-Carlo as
\begin{equation}
    \label{eq:mc_eigtaustar}
    C_{\tau^*}(\add{X}) \approx \frac{1}{mn}\left(\sum_{i \in [m]}\sum_{j \in [n]} \Hbb[S' | D\cup \tau^*_{ij}]\right) - \Hbb[S'\mid D].
\end{equation}
Assuming the dynamics are modelled with a Gaussian process, we can compute the joint Gaussian probability of the next states $S'$ \citep{rasmussen2003gaussian}.
As the entropy of a multivariate Gaussian depends only on the log-determinant of the covariance, $\log|\Sigma|$, we can tractably compute the joint entropy of the model predictions $\Hbb\left[S'\mid D\right]$ and optimize it with a zeroth order optimization algorithm.
Finally, we must calculate the entropy $\Hbb[S' | D\cup \tau^*_{ij}]$. For this, we follow a similar strategy as \citet{neiswanger2021bayesian}: since $\tau^*_i$ is a set of states given by the transition model, we can treat them as additional noiseless datapoints for our dynamics model and condition on them before computing the joint covariance matrix for $S'$.
Given this newly generalized acquisition function, we can instantiate a method of planning in order to maximize future information gained.
We give the concrete procedure for computing our acquisition function in Algorithm \ref{alg:eigtaustar}, noting that trajectories $\tau^*_{ij}$ do not depend on the query set $\add{X}$ and can be cached for various values of $\add{X}$ as long as the dataset $D$ does not change.

Our ultimate procedure, which we name \emph{Trajectory Information Planning} (TIP), is quite simple: run model-based RL using MPC as in Algorithm \ref{alg:mpc}, but set the cost function to be $C_{\tau^*}(\tau)$ instead of $C_g$ or $C_e$, and compute this cost function using Algorithm \ref{alg:eigtaustar}. At test time, we return to planning with $C_g$ as the cost function and greedily attempt to maximize returns rather than performing exploration. We can also formulate an open-loop variant of our method, oTIP, which  involves planning once and then executing the entire action sequence.
\begin{algorithm}
\caption{Computation of $C_{\tau^*}$}
\begin{algorithmic}
\STATE \textbf{Inputs:} dataset $D = \{(s_k, a_k, s'_k)\}$, query set $\add{X}$, number of start state samples $m$, number of posterior function samples $n$.
\STATE Sample $m$ start states $\{s_0^{(i)}\}_{i=1}^m \sim p_0$.
\FOR{$i \in [m]$}
\STATE Sample $n$ posterior functions $\{T'_j\}_{j=1}^n \sim P(T'\mid D)$.
\FOR{$j \in [n]$}
\STATE Set $\pi^*_j \gets \pi_{\text{MPC}}$ using $C_g$ and a singleton posterior $P(T\mid D) = \delta(T'_j)$ as in \eqref{eq:bayes_mpc}.
\STATE Compute $\tau^*_{ij}$ by executing $\pi^*_j$ on $T'_j$ starting from $s_0^{(i)}$.
\ENDFOR
\ENDFOR
\STATE Compute the joint posterior covariance $\Sigma^{S'}\mid D$ across all points in $\add{X}$.
\STATE Compute the joint posterior covariances $\Sigma^{S'}_{ij} \mid D\cup \tau_{ij}\, \forall i\in [n], j\in [m]$ across all points in $\add{X}$.
\RETURN $\log|\Sigma^{S'}| - \frac{1}{nm}\sum_{i\in[n], j\in[m]}\log|\Sigma^{S'}_{ij}|$.
\end{algorithmic}
\label{alg:eigtaustar}
\end{algorithm}

\vspace{-3mm}
\subsection{Computational Cost and Implementation Details}
\vspace{-1mm}
Though the TIP algorithm is designed for settings where samples are expensive, it is important to understand, both theoretically and practically, the computational cost of this method.
For ease of notation, we make the simplifying assumption that the planning algorithm used (in this case, iCEM from \citep{pinneri2020iCEM}) evaluates $p$ action sequences consisting of $h$ (the planning horizon) actions and that our current dataset is of size $N$.
In order to efficiently sample functions from the posterior over dynamics functions, we use the method from \citet{wilson2020efficiently}. This reduces the naive complexity of querying these functions from $O(N^3)$ to a one-time $O(N)$ cost and then $O(1)$ for additional queries. 
As we derive in Section \ref{a:computational_cost}, the computational complexity of one TIP planning iteration is $O\left(nm\left(\left(N + H\right)^3 + ph\left(N + H\right)^2\right)\right)$. The two asymptotically expensive operations are (1) computing the Cholesky decompositions of the $nm$ kernel matrices for datasets $D\cup \tau^*_{ij}$ and (2) solving the triangular systems using the cached Cholesky decompositions in order to compute the covariance matrices $\Sigma_{ij}^{S'}\mid D\cup \tau^*_{ij}$ for each of the $p$ action sequences used by the planning algorithm.

However, our implementation choices mean that in practice these operations are not the most expensive step. 
The covariance matrix computations, which are the theoretical bottleneck, are implemented in JAX \citep{jax2018github}, allowing them to be compiled to much faster machine code and vectorized across large batches of queries. In fact, the most expensive operation in practice is planning on the sampled transition functions $T'_i$ to sample optimal trajectories $\tau^*_{ij}$. This is due to the fact that in practice $p$ is large and we implemented the planner in NumPy \citep{harris2020array} so it cannot be compiled together with the Tensorflow \citep{tensorflow2015-whitepaper} code from \citet{wilson2020efficiently}, 
which is used for predicting which states will be visited for the planner. We give further information on the implementation in Section \ref{a:implementation_details}.

 \vspace{-3  mm}
\section{Experiments}
\label{sec:experiments}
\vspace{-2mm}

\begin{wrapfigure}{r}{0.45\textwidth}
    \vspace{-8mm}
    \centering
    \includegraphics[width=0.43\textwidth]{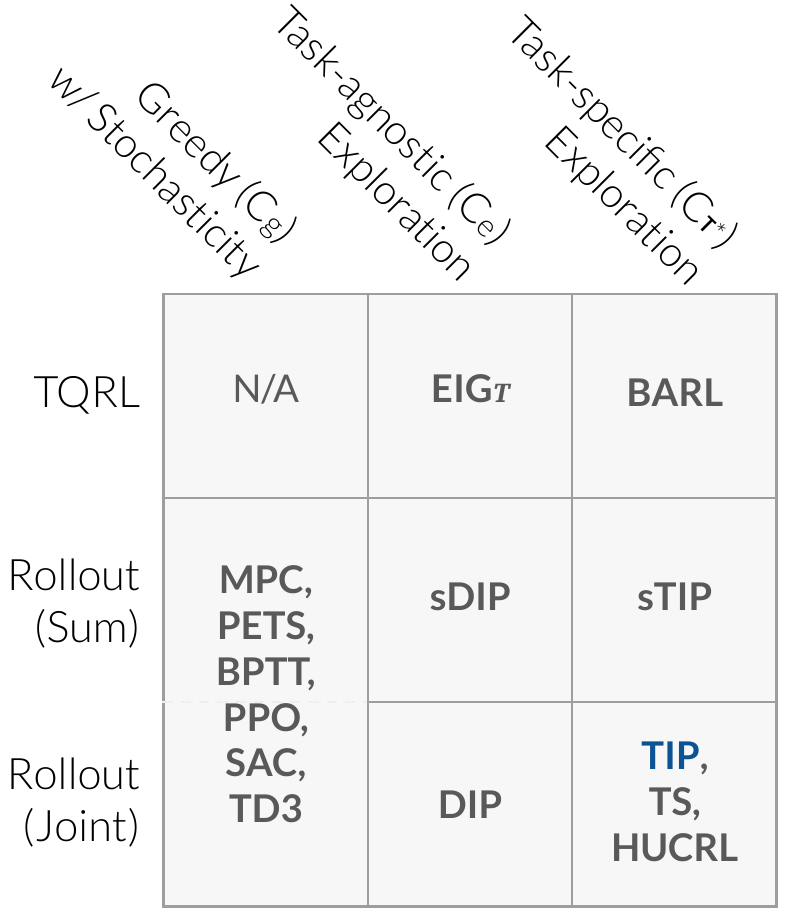}
    \vspace{-1mm}
    \caption{\small 
    Our comparison methods can be broken down by the type of cost function used and how the methods do or do not handle sequential acquisition of information.
As $C_g$ is a sum, it naturally handles future timesteps jointly. For the other information quantities, it is possible to upper-bound information acquired by summing each separate mutual information, or to compute them jointly.
    }
    \label{fig:threebythree}
    \vspace{-6mm}
\end{wrapfigure}

The aim of our development of the TIP algorithm and the $C_{\tau^*}$ acquisition function for RL is to reduce the sample complexity of learning effective policies in continuous MDPs given limited access to expensive dynamics. In this section we demonstrate the effectiveness of TIP in quickly learning a good policy by comparing against a variety of state-of-the-art reinforcement learning algorithms and strong baselines (including some that use the TQRL setting from \citep{mehta2022barl}, which is also known as RL with a generative model in \citet{kakade2003sample} and other works \citep{azar2013minimax, pmlr-v125-agarwal20b}).

In particular, we compare the average return across five evaluation episodes across five random seeds of each algorithm on five closed-loop control problems. \add{For sample complexity we assess the median amount of data taken by each algorithm to `solve' the problem across five seeds with the threshold performance given by an MPC controller using the ground truth dynamics.}
We evaluate the open-loop variant of our method, oTIP, against three comparison methods on three control problems suitable for open-loop control. In particular, to be suitable for open-loop control, the problem cannot be dynamically unstable (as Pendulum and Cartpole famously are) and must have a relatively short control horizon and fixed start state. Here too, we assess the average return as open-loop trials are conducted as well as the number of timesteps required to achieve `solved' performance.

\vspace{-2mm}
\paragraph{Comparison Methods}
We use several model-based and model-free comparison methods in this work.
We compare to several published model-based methods. These include {\bf PETS} \citep{chua_pets}, as implemented by \citet{Pineda2021MBRL}, which uses a probabilistic ensemble of neural networks and CEM over particle samples to do MPC.
We also compare against three model-based techniques from the HUCRL \citep{curi_hucrl} implementation: {\bf HUCRL} itself, which relies on hallucinating dynamics perturbations as a way of realizing an upper confidence bound on the policy, model-based Thompson Sampling ({\bf TS}), which samples from the posterior over models and chooses optimal actions for that sample, and a greedy model-based neural network method relying on backpropagation through time ({\bf BPTT}). \add{We also compare against the Free Energy of the Expected Future method from \citet{tschantz2020reinforcement}, which treats directed exploration as a process of actively collecting information for inference on a reward-biased generative model. A further comparison is with Receding Horizon Curiosity (RHC) \citep{schultheis2020receding}, which does online Bayesian system identification over a linear model in order to quickly find a model of the environment dynamics.}
Our model-free comparison methods, Soft Actor-Critic ({\bf SAC}) \citep{haarnoja2018soft}, an actor-critic method that uses an entropy bonus over the policy to encourage more exploration, \add{and two others (TD3 and PPO), are in the appendix.}

Finally, we compare against various ablations of the proposed method. These vary across two axes as described in Figure \ref{fig:threebythree}: the cost function they use and how they handle sequential queries. Besides these differences, they use the same GP model and iCEM planning algorithm with the same hyperparameters, so they are truly comparable methods.
The three cost functions used are $C_g$, $C_{\tau^*}$, and $C_e$. {\bf DIP}, {\bf oDIP}, and
\textbf{EIG$_T$} all use $C_e$ but compute, respectively, the expected joint entropy of the action sequence,
the sum of the pointwise entropies of the action sequence, and the individual pointwise entropies in the TQRL setting. \add{These methods are very similar in spirit to \citep{shyam_max, pmlr-v97-pathak19a} in that they plan for future information gain about the dynamics, but we chose to compare in a way that controls for difference in the model and planning algorithm.}
{\bf MPC} uses $C_g$ and is very close to the method in \citep{pmlr-v84-kamthe18a}. Like TIP, {\bf BARL} \citep{mehta2022barl} and {\bf sTIP} use $C_{\tau^*}$.
BARL operates in the TQRL setting and can therefore use the simpler $\EIG_{\tau^*}$ acquisition function.
sTIP investigates the use of $-\sum_{s_i, a_i \in S}\EIG_{\tau^*}(s_i, a_i)$ as a cost function for planning. This computes individual information gains for each future observation without accounting for the information they may have in common and is therefore an overestimate of the joint information gain.
In the open-loop setting we compare against {\bf oDIP} and {\bf oMPC}, the open-loop variants of DIP and MPC, and Bayesian optimization ({\bf BO}) as implemented by \citet{scikit-learn}. oDIP plans an action sequence to minimize the joint $C_e$ and executes the actions found for each open-loop trial, while oMPC does the same thing using $C_g$. We give additional details on the comparison methods in Section~\ref{a:comparison_methods}.

\begin{table}[t]
\centering
\resizebox{\columnwidth}{!}{\begin{tabular}{l|cccccccccc|cc}
    \toprule
    Environment
    & TIP & sTIP & DIP & MPC & PETS &  SAC & \add{FEEF} & \add{RHC} & HUCRL & TS & BARL & $\EIG_T$\\
\midrule
Pendulum & {\bf 21} & 36 & 36 & 46 & 5.6k & 7k & \add{800} & \add{>40k} & >50k & >50k & {\bf 21} & 56\\
Cartpole & 131 & 141 & 161 & 201 & 1.63k & 32k & \add{>2.5k} & \add{>5k} & >6k & >6k & {\bf 111} & 121\\
    $\beta$ Tracking & {\bf 46} & 76 & 276 & 76 & 330 & 12k & \add{300} & \add{>3k} & 480 & 420 & 186 & >1k\\
    $\beta$ + Rotation & {\bf 201} & >500 & >500 & >500 & 400 & 30k &\add{>2k} & \add{>2k} & >5k & >5k & >500 & >1k\\
    Reacher & {\bf 251} & >400 & >1k & 751 & 700 & 23k & \add{>5k} & \add{1.5k} & 6.6k & 4.5k & {\bf 251} & >1.5k\\
        \bottomrule
    \end{tabular}
}
\vspace{1mm}
\caption{\small
    {\bf Sample Complexity:} Median number of samples across five seeds required to reach `solved' performance, averaged across five trials.
We determine `solved' performance by
running an MPC policy (similar to the one used for evaluation) on the ground truth dynamics
    to predict actions. We record $>n$ when the median run is unable to solve the problem by the end of training after collecting $n$ datapoints. The methods in the rightmost section operate in the TQRL setting and therefore have more flexible access to the MDP dynamics for data collection. The full set of methods are shown in Section \ref{a:additional_results} as well as boxplots depicting the data in Figure~\ref{fig:boxplots}.
}
\label{tab:sample_complexity}
\vspace{-9mm}
\end{table}

\vspace{-3mm}
\paragraph{Control Problems}
Our closed loop control problems are the standard underactuated {\bf Pendulum} swing-up task (Pendulum-v0 from \citet{gym}) with 2D states and 1D actions, a {\bf Cartpole} swing-up task with a sparse reward function, 2D s, and 1D actions,
a 2-DOF robot arm control problem where the end effector is moved to a goal ({\bf Reacher}-v2 from \citet{gym}) with 10D states and 2D actions, a simplified {\bf $\beta$ Tracking} problem from plasma control \citep{char2019offline, mehta2021neural} \add{(similar in design but not identical to the one from \citet{mehta2022barl})} trained using  with 4D states and 1D actions, and a more complicated problem in plasma control where {\bf $\beta$ + Rotation} are tracked with 10D states and 2D actions.
Our open loop control problems are a navigation problem with hazards ({\bf Lava Path}) from \citep{mehta2022barl} and two regulation problems with different {\bf Nonlinear Gain} functions. The Lava Path problem has 4D states and 2D actions and the nonlinear regulation problems have 2D states and 2D actions.
Full details on these problems are available in Section \ref{a:control_problems}.

\begin{figure}[t]
    \centering
    \hspace{-3mm}\includegraphics[width=0.92\linewidth]{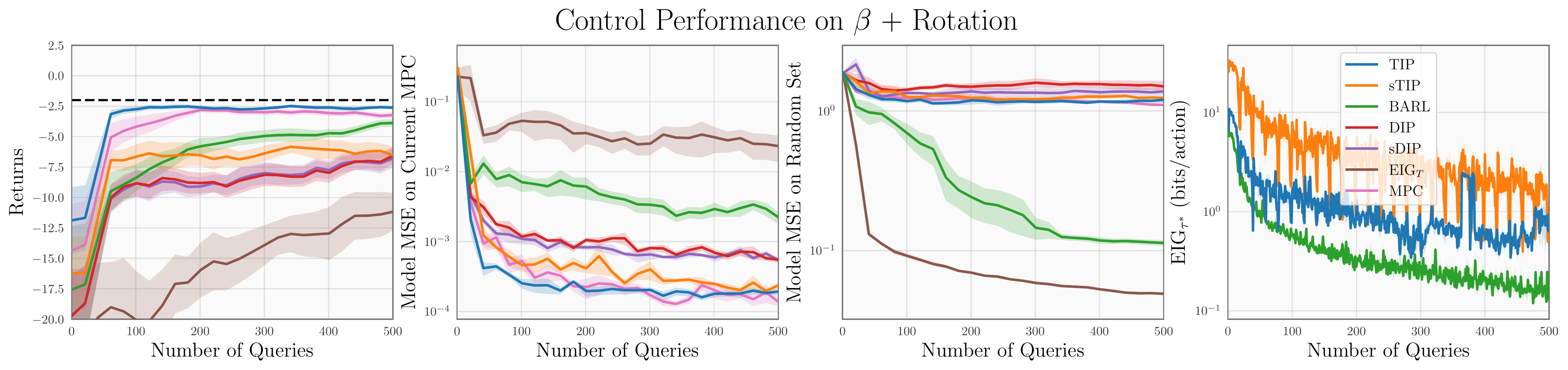}\\
    \hspace{-3mm}\includegraphics[width=0.92\linewidth]{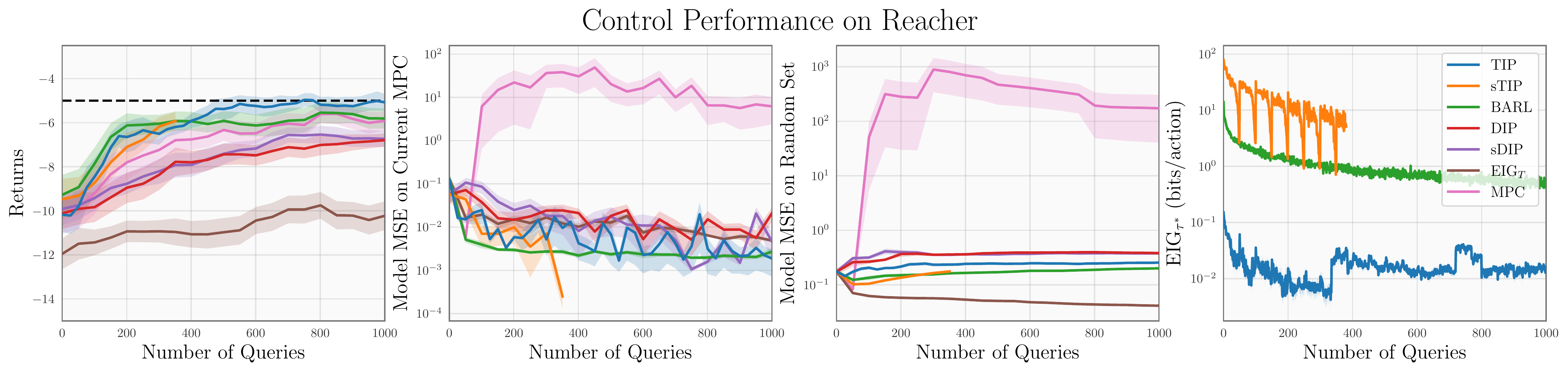}\\
    \hspace{-3mm}\includegraphics[width=0.92\linewidth]{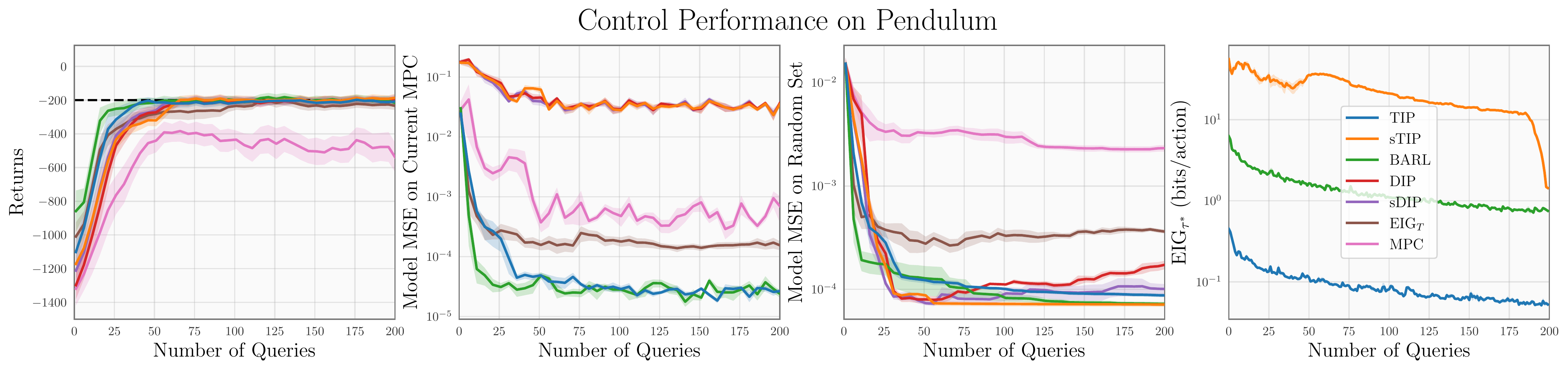}
    \vspace{-2mm}
    \caption{\small
    {\bf Control and Modeling Details for TIP and Ablations.} Column 1: Learning curves for our ablation methods, all of which use the same planner and model. Column 2: Dynamics model accuracy on the points used by the planner to choose actions during MPC. Column 3: Dynamics model accuracy on a uniformly random test set in $\tilde{\Statespace}$. Column 4: $\EIG_{\tau^*}$ values normalized by the number of actions planned. sTIP was truncated on Reacher as it exceeded the wall time budget.}
    \label{fig:beta_rotation_reacher_plots}
    \vspace{-3mm}
\end{figure}

\begin{wraptable}{r}{0.5\textwidth}
\centering
\resizebox{0.5\columnwidth}{!}{\begin{tabular}{l|cccc}
\toprule
Environment & oTIP & oMPC & oDIP & BO \\
\midrule
Nonlinear Gain 1& {\bf 41} & 91 & 51 & 210\\
Nonlinear Gain 2& {\bf 51} & 61 & >200 & 60\\
Lava Path & {\bf 41} & 101 & 101 & >2k\\
\bottomrule
\end{tabular}
}
\caption{
    \small
    {\bf Open Loop Sample Complexity:} Median number of samples
required to reach `solved' performance, averaged across five trials.
We determine `solved' performance by
running an MPC policy
on the ground truth dynamics
    to predict actions. We record $>n$ when the median run is unable to solve the problem by the end of training after collecting $n$ datapoints.
}
\vspace{-1mm}
\label{tab:open_loop}
\end{wraptable}
\vspace{-3mm}
\paragraph{Results}
As can be seen in Table \ref{tab:sample_complexity}, TIP is able to reach solved performance more quickly across the board than the model-based and model-free external baselines, often using a fraction or even orders of magnitude less data than other methods.
For many of our ablation methods we see failures to solve some of the problems even though the model is demonstrated by TIP to be able to sufficiently predict the dynamics.
This is especially apparent on the harder plasma control environment, $\beta$+Rotation, where TIP is the only method using our GP which is able to solve the problem.
We believe that this is because the data acquired through exploration by the ablation methods is less useful for control than the data TIP collects.
This is underscored by the second column of Figure \ref{fig:beta_rotation_reacher_plots}, where it is clear that TIP achieves the lowest modeling error on the points actually needed during the execution of the policy but not on the uniform test set.
In particular we find it interesting that TIP outperforms BARL on the $\beta$ + Rotation environment, as BARL should in principle have a strictly stronger access to the problem and is optimizing the same quantity with fewer constraints.
We hypothesis that this may be due to the fact that BARL optimizes the acquisition function $\EIG_{\tau^*}$ by simply uniformly sampling a set of points and choosing the one that evaluates to the largest value. Our more sophisticated optimization algorithm and forced initialization at the start state distribution seems to allow us to collect more information in this case. 
This interpretation is bolstered by the fact that on the problems where TIP outperforms BARL, we see that TIP is actually collecting more information per action than BARL as evidenced by larger $\EIG$ values. We also see clearly that there is value in computing the $C_{\tau^*}$ function rather than summing over $\EIG_{\tau^*}$ values, as TIP outperforms sTIP across the board.
Additionally, there is clear evidence for the value of task-specific exploration as the task-agnostic exploration methods ($\EIG_T$, DIP, sDIP) underperform both in returns and model error on the trajectories visited.

For the open-loop experiments (Table~\ref{tab:open_loop}), we also see strong performance from oTIP. As the model-based methods benefit from observing many model transitions for each open-loop trial, it is unsurprising that they are more sample-efficient than the BO method. Within the model-based techniques, oTIP is the most sample efficient. We believe that this is for much the same reasons as in the closed-loop case---exciting evidence that the $C_{\tau^*}$ cost function can be applied in a variety of settings.

 \section{Conclusion}
\label{sec:conclusion}

In this work, we presented and evaluated a cost function designed for intelligent, task-aware exploration. Using this cost function in model-predictive control allows agents to solve continuous MDPs with far less data than comparison methods in open- and closed-loop settings.
Though the method is effective in data reduction, it is computationally expensive and relies on dynamics that are well-modeled by a GP. In future work, we aim to scale the method to higher-dimensional and more complex control environments. We also aim to apply this method in the real world. In particular, we aim to address similar plasma control problems in a small number of trials on a real tokamak. 
 \subsection*{Acknowledgements}
This work was supported in part by US Department of Energy grants under contract numbers
DE-SC0021414 and DE-AC02-09CH1146.

This work is supported by the National Science Foundation Graduate Research Fellowship Program under Grant No. DGE1745016 and DGE2140739. Any opinions, findings, and conclusions or recommendations expressed in this material are those of the author(s) and do not necessarily reflect the views of the National Science Foundation.

This work is additionally supported by NSF (\#1651565), AFOSR (FA95501910024), ARO (W911NF-21-1-0125), CZ Biohub, and Sloan Fellowship. \bibliography{refs}
\newpage

\section*{Checklist}

\begin{enumerate}

\item For all authors...
\begin{enumerate}
  \item Do the main claims made in the abstract and introduction accurately reflect the paper's contributions and scope?
    \answerYes{}
  \item Did you describe the limitations of your work?
    \answerYes{See Section~\ref{sec:conclusion}.}
  \item Did you discuss any potential negative societal impacts of your work?
    \answerNA{}
  \item Have you read the ethics review guidelines and ensured that your paper conforms to them?
    \answerYes{}
\end{enumerate}

\item If you are including theoretical results...
\begin{enumerate}
  \item Did you state the full set of assumptions of all theoretical results?
    \answerYes{Assumptions were fully stated for all derivations.}
        \item Did you include complete proofs of all theoretical results?
    \answerNA{}
\end{enumerate}

\item If you ran experiments...
\begin{enumerate}
  \item Did you include the code, data, and instructions needed to reproduce the main experimental results (either in the supplemental material or as a URL)?
    \answerYes{All code and instructions are included in supplemental material.}
  \item Did you specify all the training details (e.g., data splits, hyperparameters, how they were chosen)?
    \answerYes{All training details are specified in the paper or in the code.}
        \item Did you report error bars (e.g., with respect to the random seed after running experiments multiple times)?
    \answerYes{We show error bars computed via the standard error of the mean performance between runs.}
        \item Did you include the total amount of compute and the type of resources used (e.g., type of GPUs, internal cluster, or cloud provider)?
    \answerYes{We detail computational cost (see Sec~\ref{a:wall_times}).}
\end{enumerate}

\item If you are using existing assets (e.g., code, data, models) or curating/releasing new assets...
\begin{enumerate}
  \item If your work uses existing assets, did you cite the creators?
    \answerYes{All existing assets are cited.}
  \item Did you mention the license of the assets?
    \answerYes{All licenses are properly attributed.}
  \item Did you include any new assets either in the supplemental material or as a URL?
    \answerYes{New assets are included in supplementary material.}
  \item Did you discuss whether and how consent was obtained from people whose data you're using/curating?
    \answerNA{No human data was used.}
  \item Did you discuss whether the data you are using/curating contains personally identifiable information or offensive content?
    \answerNA{No data with personally identifiable information was used.}
\end{enumerate}

\item If you used crowdsourcing or conducted research with human subjects...
\begin{enumerate}
  \item Did you include the full text of instructions given to participants and screenshots, if applicable?
    \answerNA{}
  \item Did you describe any potential participant risks, with links to Institutional Review Board (IRB) approvals, if applicable?
    \answerNA{}
  \item Did you include the estimated hourly wage paid to participants and the total amount spent on participant compensation?
    \answerNA{}
\end{enumerate}

\end{enumerate}

 \newpage
\appendix

\section{Implementation Details}
\label{a:implementation_details}

\subsection{Derivation of Computational Cost}
\label{a:computational_cost}
In this section, we derive the computational complexity of the TIP algorithm. For simplicity, we focus on a single TIP planning iteration as might be done at each replanning in closed-loop control or at the start of a trial in open-loop control. In order to keep the analysis general, we assume that the chosen planning algorithm requires $p$ accesses to the model where $h$ actions are sequentially executed, giving $ph$ total queries per planner execution. We also assume that the numbers of inducing points and basis functions used in our GP posterior function sampling are constant as are the Monte Carlo hyperparameters $m, n$.

The TIP algorithm consists of the following major operations:
\begin{itemize}
    \item Sample $T'_i$ for $j \in [m]$: $O(nmN)$ total cost from sampling algorithm, where $N$ is the dataset size.
    \item Sample $\tau^*_{ij}$ for $i \in [n], j\in [m]$: $phHnm$ total cost from running the planner ($ph$ posterior function queries) $H$ times for each sampled $\tau^*$, where $H$ is the MDP horizon.
    \item Compute Cholesky decomposition for each $D \cup \tau^*_{ij}$. This takes a total of $O(nm(N + H)^3)$ operations as the augmented dataset is of size $N + H$ and Cholesky decompositions are $O(d^3)$ in the matrix size $d$.
    \item Compute posterior covariance $\Sigma^{S'}\mid D\cup \tau_{ij}$ for all $\tau_{ij}$. This involves several matrix operations but the most computationally intensive is solving $h$ triangular systems of size $(N + H) \times (N + H)$, which each take $O((N + H)^2)$ time. So the total computation here is $O(pnmh(N + H)^2)$.
    \item Compute determinants of covariance matrices for each of $p$ queries and $nm$ augmented datasets $D\cup\tau_{ij}$. Each of these operations is over a matrix of size $h\times h$ and therefore costs $O(h^3)$. So the total cost is $O(pnmh^3)$.
\end{itemize}
Summing these costs gives $O(nmN + phHnm + nm(N + H)^3 + pnmh(N + H)^2 + pnmh^3)$. Clearly the third term dominates the first, the fourth dominates the second, and since $H > h$, the fourth dominates the fifth. So, the computational cost can be summarized as 
$O\left(nm\left((N + H)^3 + ph(N + H)^2\right)\right)$.

 \subsection{Wall Times}
\label{a:wall_times}
Though TIP and oTIP are designed for applications where samples are expensive and computation is relatively inexpensive, we present in this section data on the running time of these methods. We ran all experiments on a shared research cluster available to us on large machines with hundreds of GB of memory and between 24 and 88 CPU cores. In general our implementation did not make use of more than 20 CPU cores concurrently. In Table \ref{tab:runtime}, we give the running time of the phases of the TIP algorithm. We note that the bulk of the computation in the planning procedure actually goes towards the just-in-time compilation of the JAX code that computes the cost function $C_{\tau^*}$ on sampled future trajectories. In order to allow for this compilation cost, we modified the iCEM algorithm from \citep{pinneri2020iCEM} to take fixed batch sizes as the compilation (e.g. for the $\beta$ tracking problem) takes approximately 90\% of the time required for planning. Unfortunately this compilation process must be repeated at every iteration due to the limitations of the JAX compiler. We believe that a similarly JIT-compiled implementation of the planning algorithm for sampling $\tau^*$ on posterior samples could lead to a substantial speedup and a more flexible compiler could do more still. 
\begin{table}[]
    \centering
    \begin{tabular}{l|ccccc}
        \toprule
        Control Problem & Pendulum & Cartpole & $\beta$ Tracking & $\beta$ + Rotation & Reacher \\
        \midrule
        Sample $\tau^*$ $mn$ times & 24 & 31 & 7 & 25 & 130\\
        Plan actions that minimize $C_{\tau^*}$ & 16 & 15 & 15 & 50 & 295\\
        Total for TIP Iteration & 40 & 46 & 22 & 75 & 425\\
        \midrule
        Evaluation for one episode & 5-20 & 2-10 & 2-5 & 3 - 18 & 100-500\\\bottomrule
    \end{tabular}
    \caption{Runtime in seconds for the phases of the TIP algorithm on all problems when run on the authors' CPU machines. The ranges given show the runtime for the operation at the beginning and at the end of training, as some operations run longer as more data is added.}
    \label{tab:runtime}
\end{table}

\subsection{GP Model Details}
\label{a:gp_model}

For all of our experiments, we use a squared exponential kernel with automatic relevance determination~\citep{mackay1994bayesian, neal1995bayesian}. The parameters of the kernel were estimated by maximizing the likelihood of the parameters after marginalizing over the posterior GP \citep{williams1996gaussian}.

To optimize the transition function, we simply sampled a set of points from the domain, evaluated the acquisition function, and chose the maximum of the set. This set was chosed uniformly for every problem but $\beta$ + Rotation and Reacher, for which we chose a random subset of $\cup_i\cup_j \tau^*_{ij}$ (the posterior samples of the optimal trajectory) since the space of samples is 10-dimensional and uniform random sampling will not get good coverage of interesting regions of the state space.

\subsection{Cost Function Details}
\label{a:cost_function}
We set $n = 15$ and $m = 1$ for our Monte Carlo estimate of the cost function for each problem.

\subsection{Details on Planning Method}
\label{a:planning}
As mentioned in the main text, we use the iCEM method from \citet{pinneri2020iCEM} with one major modification: a fixed sample batch size. This is in order to take advantage of the JIT compilation features of JAX and avoid recompiling code for each new batch size. 

In Tables \ref{tab:MPC_hypers} and \ref{tab:open_loop_hypers}, we present the hyperparameters used for the planning algorithm across each problem. The same hyperparameters were used for the TIP, MPC, $\EIG_T$, DIP, sDIP, and sTIP methods. As recommended by the original paper, we use $\beta = 3$ for the scaling exponent of the power spectrum density of sampled noise for action sequences, $\gamma = 1.25$ for the exponential decay of population size, and $\xi = 0.3$ for the amount of caching.

\begin{table}[]
\centering
\begin{tabular}{l|ccccc}
\toprule
Control Problem & Pendulum & Cartpole & $\beta$ Tracking & $\beta$ + Rotation & Reacher\\
\midrule
Number of samples & 25 & 30 & 25 & 50 &  100  \\
Number of elites & 3 & 6 & 3 & 8 & 15  \\
Planning horizon & 20 & 15 & 5 & 5 & 15 \\
Number of iCEM iterations & 3& 5 & 3 & 5 & 5\\
Replanning Period & 6 & 1 & 2 & 1 & 1 \\
\bottomrule
\end{tabular}
\vspace{2mm}
\caption{Hyperparameters used for optimization in MPC procedure for closed-loop control problems.}
\label{tab:MPC_hypers}
\end{table}

\begin{table}[]
\centering
\begin{tabular}{l|ccc}
\toprule
Control Problem & Nonlinear Gain 1 & Nonlinear Gain 2 & Lava Path\\
\midrule
Number of samples & 50 & 50 & 25   \\
Number of elites & 6 & 6 & 4  \\
Planning horizon & 10 & 10 & 20 \\
Number of iCEM iterations & 6& 8 & 6 \\
\bottomrule
\end{tabular}
\vspace{2mm}
\caption{Hyperparameters used for optimization in MPC procedure for open-loop control problems.}
\label{tab:open_loop_hypers}
\end{table}

\section{Description of Comparison Methods}
\label{a:comparison_methods}

We compare against 14 different methods across open and closed-loop problems. Of these, 7 used the same model and planning algorithm (including hyperparameters) as TIP and oTIP. {\bf DIP} and {\bf oDIP} use the cost function $C(\tau) = - \Hbb\left[T(S') \mid D \right]$ and {\bf sDIP} (summed DIP) uses the cost function $C(\tau) = - \sum_{i=0}^h \Hbb\left[T(s_i, a_i) \mid D \right]$. These are all pure exploration methods, but DIP and oDIP are more sophisticated in that they plan for future observations with a large amount of \emph{joint} information as opposed to sTIP which sums the individual information expected at each timestep. oDIP is simply the open loop variant of DIP. \textbf{EIG}$_T$ uses the same objective as sDIP but operates in the TQRL setting, querying points that approximately maximize the predictive entropy of the dynamics model. {\bf BARL} similarly operates in the TQRL setting but uses the $\EIG_{\tau^*}$ acquisition function from \citet{mehta2022barl}. We use the authors' implementation of that work for comparison.
{\bf MPC} uses $C_g$ from \eqref{eq:cost_functions} and plans to directly maximize expected rewards. This method can be seen as quite similar to \citet{pmlr-v84-kamthe18a} and a close cousin of \citet{pilco} in that it optimizes the same objective with a similar model. {\bf oMPC} is simply the open loop variant of MPC.

Besides these methods which directly compare cost functions, we include 8 additional baselines from published work. {\bf PETS} is a method given in \citet{chua_pets} which uses a similar cross-entropy based planner and a probabilistic ensemble of neural networks for an uncertainty-aware estimate of the dynamics. PETS also plans to minimize $C_g$. {\bf HUCRL} \citep{curi_hucrl} learns a policy via backpropagation through time using a hallucinated perturbation to the dynamics that maximizes discounted rewards subject to the one-step confidence interval of the dynamics. HUCRL also uses a probabilistic ensemble of neural networks. Using the same implementation we also tested Thompson Sampling ({\bf TS}), which acts optimally according to a network drawn from the posterior over models, and {\bf BPTT} which plans to minimize $C_g$ using a neural network policy and backpropagation through time. BPTT can also be viewed as a cousin of PILCO \citep{pilco} as it attempts to take stochastic gradients of the expected cost. We also compare against {\bf SAC} \citep{haarnoja2018soft}, {\bf TD3} \citep{TD3Fujimoto}, and {\bf PPO} \citep{schulman2017proximal}. SAC uses entropy bonuses to approximate Boltzmann exploration in an actor-critic framework. TD3 and PPO include various tricks for stable learning and add Ornstein-Uhlenbeck noise in order to explore.

\add{For our FEEF implementation, we took hyperparameters from the most similar comparison environments in that paper and used them for our results. We tried several values for `expl\_weight' in order to se whether we were inadequately balancing exploration and exploitation. Ultimately we saw an `expl\_weight' of 0.1 was the best value.}

\add{We used the author's implementation of RHC. RHC makes strong assumptions on the form of the reward function by assuming that all problems are regulation problems where the goal is to drive the system to a given state and keep it there (with some cost for actuation). We were able to pass the targets for all of our problems (which may change between episodes) to the RHC controller. We did a light hyperparameter search tuning the number of random Fourier features used in the Bayesian linear model in this method. Ultimately we were disappointed in the performance of RHC when applied to our problems. We believe that this might be due to its undirected uncertainty sampling objective and relatively constrained model of environment dynamics.}

\section{Description of Control Problems}
\label{a:control_problems}
\subsection{Plasma Control Problems}
\label{a:plasma_control}

The plasma control problems are based on controlling a tokamak, a toroidally shaped device for confining a thermonuclear plasma using magnetic fields. Achieving net positive energy from fusion requires confining a plasma at high enough temperature and density long enough for hydrogen isotopes to collide and fuse. However, as the temperature and density are increased, a wide variety of instabilities can occur which degrade confinement, leading to a loss of energy. Full physics simulation of tokamak plasmas requires 10s-1000s of CPU hours to simulate a single trajectory, and often require hand tuning of different parameters to achieve accurate results. Following the work of \citet{abbate2021data}, each of our plasma control problems used neural networks trained on data as the ground truth dynamics models. We used the MDSPlus tool \citep{stillerman1997mdsplus} to fetch historical discharges from the DIII-D tokamak in San Diego \citep{fenstermacher2022diii}. In total, we trained our models on 1,479 historical discharges. The data was pre-processed following the procedure outlined in \citet{abbate2021data}. We describe how each environment was constructed in more detail below.

\paragraph{$\beta$ Tracking} In this environment the goal is to adjust the total injected power (PINJ) of the neutral beams so that the normalized plasma pressure, $\beta_N$ (defined as the ratio of thermal energy in the plasma to energy in the confining magnetic fields), reaches a target value of 2\%. Reliably controlling plasmas to sustain high performance is a major goal of research efforts for fusion energy, so even this simple scenario is of interest. The ground-truth dynamics model takes in the current $\beta_N$ and PINJ, the $\beta_N$ and PINJ at some $\Delta t$ time in the past, and the PINJ at some $\Delta t$ time in the future (we assume that we have complete control over the values of PINJ at all times). Given these inputs, the model was trained to output what $\beta_N$ will be $\Delta t$ time into the future. In total, the state space is 4D and the action space is 1D. For this environment, we set $\Delta t = 200 ms$, and we specify the reward function to be the negative absolute difference between the next $\beta_N$ and the target $\beta_N = 2\%$.

\paragraph{$\beta$ + Rotation Tracking} This environment is a more complicated version of the $\beta$ tracking environment in several ways. First of all, the controller now must simultaneously track both $\beta_N$ and the core toroidal rotation of the plasma. To do so, the controller is also allowed to set the total torque injected (TINJ) of the neutral beams (DIII-D has eight neutral beam injectors at different positions around the tokamak, so it is generally possible to control both total power and total torque independently). Controlling both of these quantities simultaneously is of interest since rotation shear often results in better confinement and less chance of instabilities in the plasma \citep{Bondeson_Ward_1994, Groebner_Burrell_Seraydarian_1990}. In addition, we assume a multi-task setting where the requested targets for $\beta_N$ and rotation can be set every trajectory. Specifically, the $\beta_N$ target is drawn from $U(1.5\%, 2.5\%)$ and the rotation target is drawn from $U(25, 125)$ krad/s every trajectory. These targets are appended to the state space. 

The learned, ground-truth dynamics model is also more sophisticated here. In addition to the inputs and outputs used by the $\beta$ tracking environment model, the inputs for this model also include rotation and TINJ at times $t$, $t -\Delta t$, and $t + \Delta t$ for TINJ only. This model receives additional information about the plasma (e.g. the shape of the plasma); however, we have assumed these inputs are fixed to reasonable values in order to avoid partial observability problems. In total, the state space of this problem is 10D (targets plus current and past observations for $\beta_N$, rotation, PINJ, and TINJ) and the action space is 2D (next PINJ and TINJ settings). 
\subsection{Robotics Problems}
\label{a:robotics}
\paragraph{Pendulum} The pendulum swing-up problem is the standard one found in the OpenAI gym \citep{gym}. The state space contains the angle of the pendulum and its first derivative and action space simply the scalar torque applied by the motor on the pendulum. The challenge in this problem is that the motor doesn't have enough torque to simply rotate the pendulum up from all positions and often requires a back-and-forth swing to achieve a vertically balanced position. The reward function here penalizes deviation from an upright pole and squared torque.

\paragraph{Cartpole} The cartpole swing-up problem has 4-dimensional state (position of the cart and its velocity, angle of the pole and its angular velocity) and a 1-dimensional action (horizontal force applied to the cart). Here, the difficulty lies in translating the horizontal motion of the cart into effective torque on the pole. The reward function is a negative sigmoid function penalizing the distance betweent the tip of the pole and a centered upright goal position.

\paragraph{Reacher} The reacher problem simulates a 2-DOF robot arm aiming to move the end effector to a randomly resampled target provided. The problem requires joint angles and velocities as well as an indication of the direction of the goal, giving an 8-dimensional state space along with the 2-dimensional control space.

\section{Additional Results}
\label{a:additional_results}

Due to space constraints in the main paper, we omitted results for the methods sDIP and BPTT. The are included alongside the rest in Table \ref{tab:appendix_sample_complexity}. They are outperformed across the board by TIP.
\begin{table}[t]
\centering
\resizebox{\columnwidth}{!}{\tiny
    \begin{tabular}{l|ccccccccccccc|cc}
    \toprule
    Environment
    & TIP & sTIP & DIP & sDIP & MPC & PETS &  SAC & TD3 & PPO & FEEF & HUCRL & TS & BPTT & BARL & $\EIG_T$\\
\midrule
Pendulum & {\bf 21} & 36 & 36 & 46 & 46 & 5.6k & 7k & 26k & 14k & 800 &  >50k & >50k & >50k & {\bf 21} & 56\\
Cartpole & 131 & 141 & 161 & 141 & 201 & 1.63k & 32k & 18k & >1M & >2.5k & >6k & >6k & >6k & {\bf 111} & 121\\
    $\beta$ Tracking & {\bf 46} & 76 & 276 & 131 & 76 & 330 & 12k & 17k & 39k & 300 & 480 & 420 & 450 & 186 & >1k\\
    $\beta$ + Rotation & {\bf 201} & >500 & >500 & >500 &  >500 & 400 & 30k & >50k & >50k & >2k & >5k & >5k & >5k & >500 & >1k\\
    Reacher & {\bf 251} & >400 & >1k & >1k & 751 & 700 & 23k & 13k & >100k & >5k & 6.6k & 4.5k & 3.7k & {\bf 251} & >1.5k\\
        \bottomrule
    \end{tabular}
}
\vspace{1mm}
\caption{\small
    {\bf Sample Complexity Comparison of All Methods:} Median number of samples across 5 seeds required to reach `solved' performance, averaged across 5 trials.
We determine `solved' performance by
running an MPC policy (similar to the one used for evaluation) on the ground truth dynamics
    to predict actions. We record $>n$ when the median run is unable to solve the problem by the end of training after collecting $n$ datapoints. The methods in the rightmost section operate in the TQRL setting and therefore have more flexible access to the MDP dynamics for data collection. 
}
\label{tab:appendix_sample_complexity}
\end{table}

\begin{figure}
    \centering
    \includegraphics[width=0.3\textwidth]{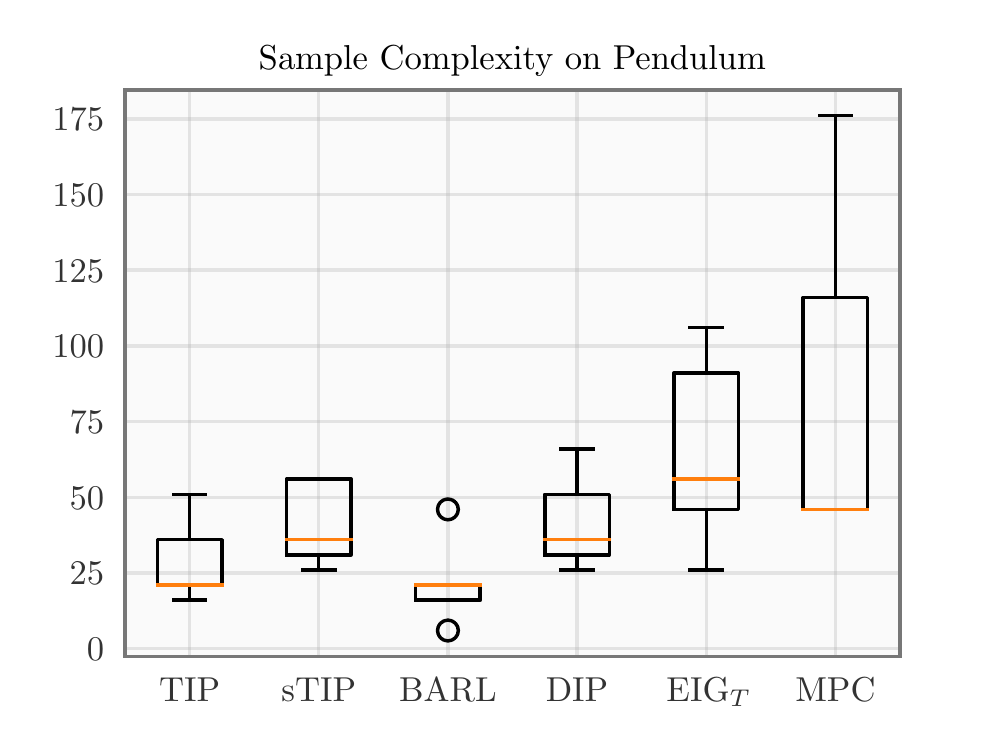}
    \includegraphics[width=0.3\textwidth]{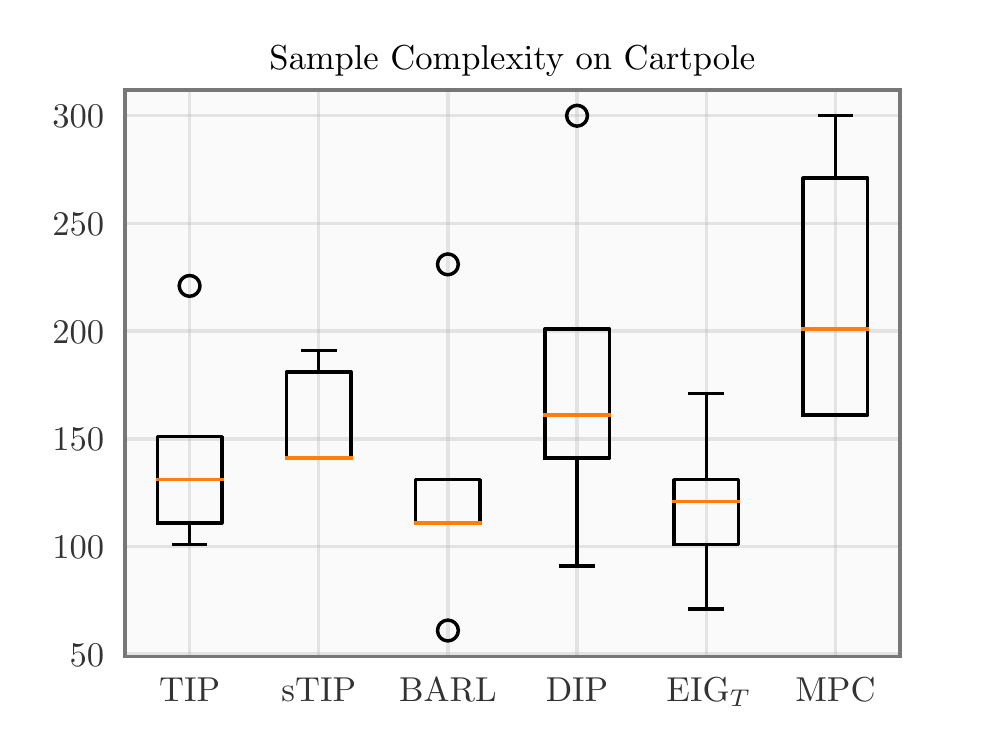}
    \includegraphics[width=0.3\textwidth]{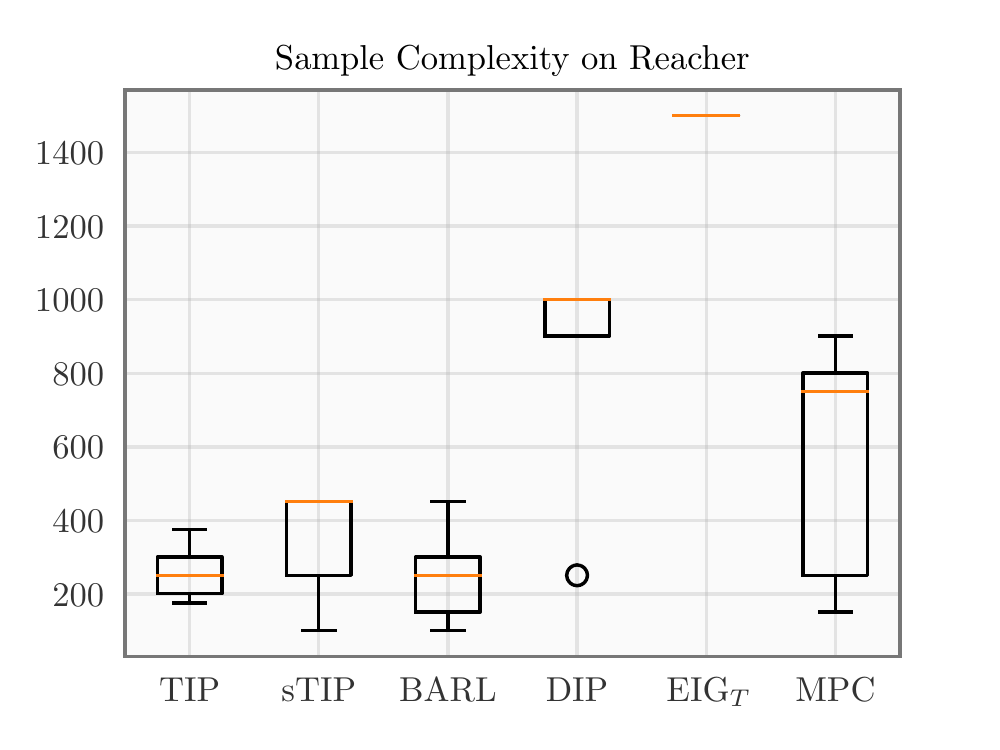}
    \includegraphics[width=0.3\textwidth]{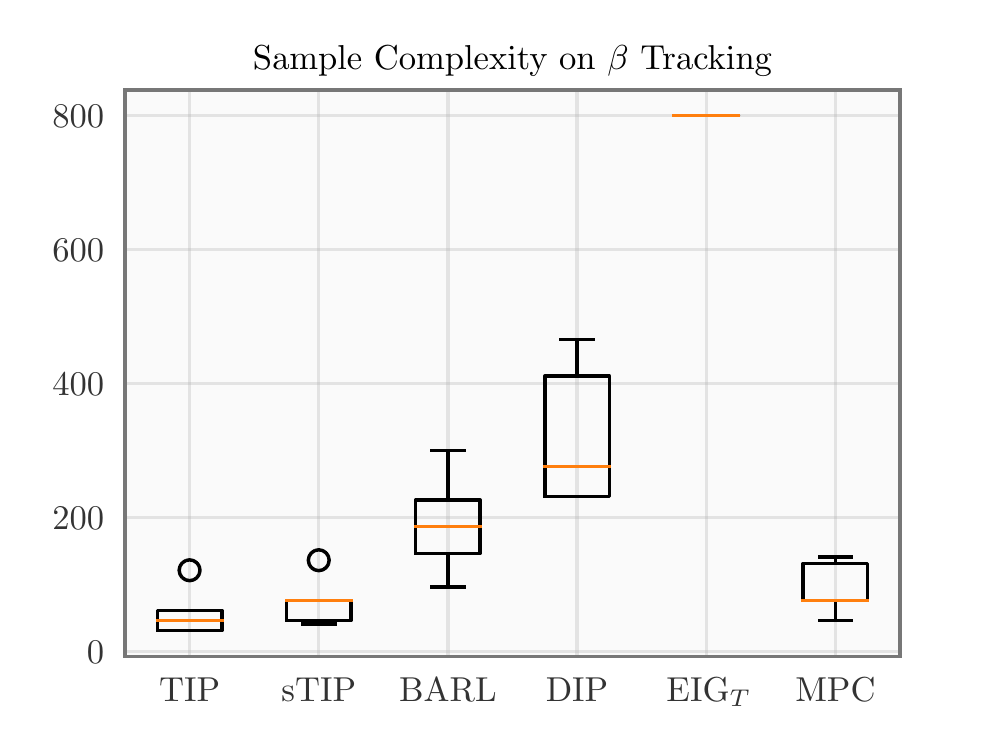}
    \includegraphics[width=0.3\textwidth]{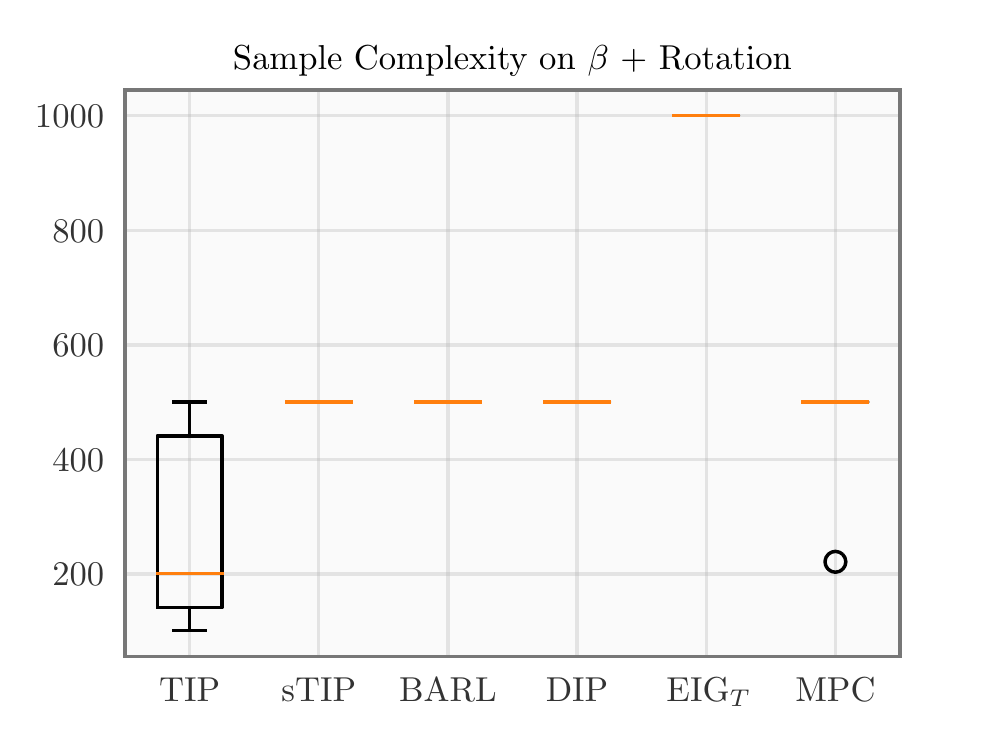}
    \caption{Box plots showing sample complexity figures across the 5 random seeds run. Each of these show for a given training run how many samples were needed to achieve the performance of an MPC controller given ground truth dynamics averaged across test episodes. We imputed the maximum number of samples for agents that failed to ever solve the problem on a given run.}
    \label{fig:boxplots}
\end{figure}

\section{Additional Related Work}
\label{a:rw}
\subsection{Bayesian Exploration Techniques}
\label{a:bayesian_exploration}
Given unlimited computation and an accurate prior, solving the Bayes-adaptive MDP \citep{ross2007bayes} gives an optimal tradeoff between exploration and exploitation by explicitly accounting for the updated beliefs that would result from future observations and planning to find actions that result in high rewards as quickly as can be managed given the current posterior. However, this is computationally expensive even in small finite MDPs and totally intractable in continuous settings.
\citet{kolter2009nearbayesian} and \citet{guez_bayes_mdp} show that even approximating these techniques can result in substantial theoretical reductions in sample complexity compared to frequentist PAC-MDP bounds as in \citet{kakade2003sample}. Another line of work \citep{dearden1998bayesian, DeardenMBBE} uses the myopic value of perfect information as a heuristic for similar Bayesian exploration in the tabular MDP setting.
Further techniques for exploration include knowledge gradient policies \citep{ryzhov2019bayesian, ryzhov2011information}, which approximate the value function of the Bayes-adaptive MDP and information-directed sampling (IDS) \citep{RussoIDS}, which takes actions based on minimizing the ratio between squared regret and information gain over dynamics. This was extended to continuous-state finite-action settings using neural networks in \citet{nikolov2018informationdirected}. 
Another very relevant recent paper \citep{ball_rp1} gives an acquisition strategy in policy space that iteratively trains a data-collection policy in the model that trades off exploration against exploitation using methods from active learning. \citet{context-achterhold21a} use techniques from BOED to efficiently calibrate a Neural Process representation of a distribution of dynamics to a particular instance, but this calibration doesn't include information about the task. A tutorial on Bayesian RL methods can be found in \citet{GhavamzadehBayesRL} for further reference.

\subsection{Gaussian Processes (GPs) in Reinforcement Learning}
\label{a:gp_rl}
There has been substantial prior work using GPs \citep{rasmussen2003gaussian} in reinforcement learning. Most well-known is PILCO \citep{pilco}, which computes approximate analytic gradients of policy parameters through the GP dynamics model while accounting for uncertainty. The original work is able to propagate the first 2 moments of the occupancy distribution through time using the GP dynamics and backpropagate gradients of the rewards to policy parameters. In \cite{wilson2020efficiently}, a method is developed for efficiently sampling functions from a GP posterior with high accuracy. One application show in their work is a method of using these samples to backpropagate gradients of rewards through time to policy paramters, which can be interpreted as a different sort of PILCO implementation. Most related to our eventual MPC-based method is \citep{pmlr-v84-kamthe18a}, which gives a principled probabilistic model-predictive control algorithm for GPs. We combine ideas from this paper, PETS \citep{chua_pets}, and the ability to sample posterior functions discussed above to give our eventual MPC component as discussed in Section \ref{sec:methods-mpc}.

\end{document}